\documentclass[11pt,a4paper]{article}
\usepackage[hyperref]{acl2021}
\usepackage{lingmacros}
\usepackage{times}
\usepackage{latexsym}
\usepackage{tipa}
\usepackage{multirow}

\usepackage{microtype}

\usepackage{soul}

\title{Not quite there yet: Combining  analogical patterns and\\ encoder-decoder networks for cognitively plausible inflection}

\author{Basilio Calderone\textsuperscript{a} \hspace{5ex} Nabil Hathout\textsuperscript{a} \hspace{5ex} Olivier Bonami\textsuperscript{b} \\
  \textsuperscript{a}CNRS, CLLE, Université de Toulouse \hspace{5ex}
 \textsuperscript{b}Université de Paris, LLF, CNRS\\  \texttt{basilio.calderone@univ-tlse2.fr}\\
  \texttt{nabil.hathout@univ-tlse2.fr}\\
  \texttt{olivier.bonami@u-paris.fr}}
 
\date{}

\aclfinaltrue

\begin{document}
\maketitle
\begin{abstract}
The paper presents four models submitted to Part 2 of the SIGMORPHON 2021 Shared Task 0, which aims at replicating human judgements on the inflection of nonce lexemes.  Our goal is to explore the usefulness of combining pre-compiled analogical patterns with an encoder-decoder architecture.  Two models are designed using such patterns either in the input or the output of the network.  Two extra models controlled for the role of raw similarity of nonce inflected forms to existing inflected forms in the same paradigm cell, and the role of the type frequency of analogical patterns.  Our strategy is entirely endogenous in the sense that the models appealing solely to the data provided by the SIGMORPHON organisers, without using external resources.  Our model 2 ranks second among all submitted systems, suggesting that the inclusion of analogical patterns in the network architecture is useful in mimicking speakers' predictions.
\end{abstract}

\section{Introduction}
\label{sec:intro}

Psycho-computational experiments deal with the capability of the computational models of language to capture and mimic the behavioural responses of speakers exposed to the same data stimuli. In phonology, and in morphology, a large number of computational models (both symbolic and sub-symbolic) have been tested on various experimental data in order to evaluate their capacity to simulate linguistic behavior \citep{rumelhart1986lpt,Nosofsky90,albright-hayes-2003-rules,Hahn_Bailey:2005,hay_pierrehumbert_beckman_2004,Albright:2009}. 

More recently, \citet{kirov-cotterell-2018-recurrent} argue that a neural encoder-decoder network (ED) can perform morphological inflection tasks in a cognitively valid manner. In particular, the authors claim that, in a \textit{wug} test protocol, ED's outputs significantly correlate with human judgements for nonce verbs supporting the assumption that the model learns representations of specific knowledge and involves cognitive mechanisms that are known to underlie language processing in the speakers.  However, whether and how such models are able to mimic the human behaviour of subjects exposed to the same stimuli is still an open question \citep{corkery2019yet}. 
Part 2 of Shared Task 0 addresses this same issue.  More specifically, it adopts the experimental approach of \citet{albright-hayes-2003-rules}.  The task is to design models which predict the inflected forms of a set of nonce verbs in a given language and that have output scores of the predictions that correlate with human judgements. 

In this paper we report on a series of experiments that address the shared task by exploring whether pre-learned formal analogies between strings can be usefully combined with an ED architecture to alleviate some limitations of the application of an ED architecture to raw forms. We present two types of models using analogical patterns in the input (M1) vs. in the output (M2) of an ED network, and compare their performance to that of two baselines that focus respectively on the phonotactic typicality of outputs (M3) and the type frequency of alternations (M4). We report good performance for the M2 architecture and quite poor performance for M1. 

In Section~\ref{sec:data}, we shortly present the data provided by the organisers.  Section~\ref{sec:patterns} focuses on the analogical patterns we use in our models.  The models architecture, the training parameters and the results are reported in section \ref{sec:combining}.  We then discuss the results in Section~\ref{sec:discussion} and draw conclusions (Section~\ref{sec:conclusion}). 

\section{Data and goal}
\label{sec:data}

The linguistic data provided by the organisers are inflected verb forms in English (ENG), German (DEU) and Dutch (NLD). For each language, four datasets are released: (a) a training dataset, (b) a development dataset of attested verb forms, (c) a development dataset of \textit{wug} forms which includes human judgements, and (d) a test dataset of \textit{wug} forms without the human judgements.  The goal of the Shared Task is to assign a  score to each \textit{wug} form in (d) that correlates as closely as possible with the human judgements (see Section~\ref{sec:results} for more details on the evaluation process).  

The entries of the datasets include lemma/form pair and a UniMorph  \citep{kirov-etal-2018-unimorph} tag (UT)  specifying the part of speech and the paradigm cell of the form.  The pairs are provided as written forms (orthographically) in datasets (a) and (b) and in IPA (phonologically) in all four datasets.

Table \ref{data_analysis} summarizes the size of the training data (a), its phonological  make-up, the number of morphosyntactic tags and the proportion ($\%$) of syncretism, i.e.\ of forms that fill two or more paradigm cells of the same lexeme. 
Although the number of phonemes is substantially similar in the three languages, the datasets differ in the number of entries (twice as many entries in DEU as in ENG) and  number of cells.  The three datasets have a comparable amount of syncretism.
\begin{table}[ht] 
\centering
\begin{tabular}[t]{|l|rrr|} 
\hline
 & ENG & DEU & NLD \\
\hline
entries & $41~658$ & $100~011$ & $74~176$ \\
phonemes & $43$ & $44$ & $39$ \\
UTs & $11$ & $29$ & $7$ \\
syncretism $\%$ & $53$ & $50$ & $42$  \\
\hline
\end{tabular}
\caption{Training data}
\label{data_analysis}
\end{table}

\section{Analogical Patterns}
\label{sec:patterns}

Three of our model architectures rely on alternation patterns (APs) describing the formal relationship between two word-forms.  An AP is a pairing of two word-forms patterns (WPs) with shared variables over substrings which represent word parts that are common between the two forms.   For example, the two German word-forms \emph{\textipa{anSpi\textlengthmark{}l@n}} `to allude to' and \emph{\textipa{anSpi\textlengthmark{}l@nt}} `alluded to' are related by the pattern (\texttt{an$X$\textipa{@}n}, \texttt{an$X$\textipa{@}nt}) where the variable $X$ represents \texttt{\textipa{S}p\textipa{i\textlengthmark{}}l}.

The number of different APs satisfied by a pair of forms is typically large.  For instance,  there are $256$ ($2^8$) distinct patterns relating \emph{\textipa{anSpi\textlengthmark{}l@n}} and \emph{\textipa{anSpi\textlengthmark{}l@nt}}, some of which are shown in (\ref{ex:patterns-1}), where italic capital letters represent variables over strings.
\enumsentence{
  \label{ex:patterns-1}
  \evnup[5pt]{%
    \begin{tabular}{lll}
    \multicolumn{1}{c}{WP$_1$} &  \multicolumn{1}{c}{WP$_2$}\\
    \cline{1-2}
      \texttt{an\textipa{S}p\textipa{i\textlengthmark{}}l\textipa{@}n} & \texttt{an\textipa{S}p\textipa{i\textlengthmark{}}l\textipa{@}nt} &  $\leftarrow$ TAP\\
      \texttt{$X$n\textipa{S}p\textipa{i\textlengthmark{}}l\textipa{@}n} & \texttt{$X$n\textipa{S}p\textipa{i\textlengthmark{}}l\textipa{@}nt} &  \\
      \texttt{a$X$\textipa{S}p\textipa{i\textlengthmark{}}l\textipa{@}n} & \texttt{a$X$\textipa{S}p\textipa{i\textlengthmark{}}l\textipa{@}nt} &  \\
      \texttt{an$X$p\textipa{i\textlengthmark{}}l\textipa{@}nt} &
      \texttt{an$X$p\textipa{i\textlengthmark{}}l\textipa{@}n} & \\ \texttt{a$X$\textipa{S}$Y$\textipa{i\textlengthmark{}}$Z$\textipa{@}$T$} & \texttt{a$X$\textipa{S}$Y$\textipa{i\textlengthmark{}}$Z$\textipa{@}$T$t} &  \\
      \texttt{$X$n$Y$p$Z$l$T$n} & \texttt{$X$n$Y$p$Z$l$T$nt} &  \\
      \texttt{$X$\textipa{@}n} & \texttt{$X$\textipa{@}nt} & $\leftarrow$ FAP \\
      \texttt{a$X$} & \texttt{a$X$t} &  \\
      \texttt{$X$n$Y$} & \texttt{$X$n$Y$t} &  \\
      \texttt{$X$n} & \texttt{$X$nt} &  \\
      \texttt{$X$} & \texttt{$X$t} & $\leftarrow$ BAP
    \end{tabular}%
  }
}

Henceforth we will note patterns using the ‘\texttt{+}’ symbol as a general notation for variables, and rely implicitly on order to match variables in alternations. Hence e.g.\ the AP (\texttt{$X$n$Y$p$Z$l$T$n}, \texttt{$X$n$Y$p$Z$l$T$nt}) will be noted \texttt{+n+p+l+n/+n+p+l+nt}.

Most of the patterns in~(\ref{ex:patterns-1}) are of little morphological interest. This is in particular the case of the  trivial alternation pattern (TAP) which just records the two strings without making any generalization over common elements. A broad alternation pattern (BAP) is an optimal pattern that can be inferred by pairwise alignment of the two forms, without taking into account the situation in the rest of the paradigm. In principle there can be more than one BAP for a pair of form, although that rarely happens in practice. This type of pattern is of crucial interest to the study of the implicative structure of paradigms \citep{Ackerman09, Ackerman13} and the induction of inflection classes \citep{Beniamine17}, but does not lead to the identification of affixes familiar from typical grammatical descriptions.  For that purpose, multiple alignments across the paradigms are necessary \citep{Beniamine21}, and lead to what we call a fine alternation pattern (FAP): here \emph{\textipa{@n}} is the infinitive suffix and \emph{\textipa{@nt}} is the present participle suffix.

The crucial intuition behind the determination of FAPs is that they identify recurrent partials \citep{Hockett47} across both paradigms and lexemes. For instance, the FAP in  (\ref{ex:patterns-1}) is motivated by the fact that the substrings \emph{\textipa{an}} and \emph{\textipa{Spi\textlengthmark{}l}} are shared across all pairs of paradigm cells of \emph{anspielen} (\ref{ex:flex-anspielen-1}), while the substrings \textipa{\em @n} and \textipa{\em @nt} recur in many (infinitive, present participle) pairs across lexemes (\ref{ex:flex-other-verbs-1}).  

\enumsentence{
  \label{ex:flex-anspielen-1}
  \evnup[5pt]{%
    \begin{tabular}{lll}
      \textipa{anSpi\textlengthmark{}l\textbf{@n}}&\textipa{Spi\textlengthmark{}l\textbf{@t}+an}&V;IND;PST;2;PL\\
      \textipa{anSpi\textlengthmark{}l\textbf{@n}}&\textipa{Spi\textlengthmark{}l\textbf{t@}+an}&V;SBJV;PST;1;SG\\
      \textipa{anSpi\textlengthmark{}l\textbf{@n}}&\textipa{Spi\textlengthmark{}l\textbf{t}+an}&V;IMP;2;PL\\
      \textipa{anSpi\textlengthmark{}l\textbf{@n}}&\textipa{Spi\textlengthmark{}l\textbf{@}+an}&V;IMP;2;SG\\
      \textipa{anSpi\textlengthmark{}l\textbf{@n}}&\textipa{anSpi\textlengthmark{}l\textbf{@st}}&V;SBJV;PST;2;SG\\
      \textipa{anSpi\textlengthmark{}l\textbf{@n}}&\textipa{anSpi\textlengthmark{}l\textbf{st}}&V;IND;PRS;2;SG
    \end{tabular}%
  }
}

\enumsentence{
  \label{ex:flex-other-verbs-1}
  \evnup[5pt]{%
    \begin{tabular}{l@{~~~}l@{~~}l}
      \textipa{tari\textlengthmark{}fi\textlengthmark{}r\textbf{@n}} & \textipa{tari\textlengthmark{}fi\textlengthmark{}r\textbf{@nt}} & V.PTCP;PRS \\
      `to tariff' &&\\
      \textipa{tari\textlengthmark{}fi\textlengthmark{}r\textbf{@n}} & \textipa{tari\textlengthmark{}fi\textlengthmark{}r\textbf{t@n}} & V;SBJV;PST;3;PL \\
      \textipa{ast\textbf{@n}} & \textipa{ast\textbf{@nt}} & V.PTCP;PRS  \\
      `to lug' &&\\
      \textipa{ast\textbf{@n}} & \textipa{ast\textbf{@t@t}} & V;SBJV;PST;2;PL  \\
      \textipa{va\textsubarch{i}n\textbf{@n}} & \textipa{va\textsubarch{i}n\textbf{@nt}} & V.PTCP;PRS \\
      `to cry' &&\\
      \textipa{va\textsubarch{i}n\textbf{@n}} & \textipa{va\textsubarch{i}n\textbf{t}} & V;IND;PRS;3;SG  \\
      \textipa{tsErStra\textsubarch{i}t\textbf{@n}} & \textipa{tsErStra\textsubarch{i}t\textbf{@nt}} & V.PTCP;PRS \\
      \multicolumn{2}{l}{`to disagree'} &\\
      \textipa{tsErStra\textsubarch{i}t\textbf{@n}} & \textipa{tsErStra\textsubarch{i}t\textbf{@st}} & V;SBJV;PST;2;SG 
    \end{tabular}%
  }
}

In this paper, we rely on an algorithm for inferring BAPs and FAPs initially developed to create Glawinette \citep{hathout2020.lrec-glawinette}.  Glawinette is a French derivational lexicon created from the definitions of the GLAWI machine readable dictionary \citep{sajous2015.glawi,hathout2016.lrec-glawi}.  Glawinette provides a description of derivational morphology by means of morphological families and derivational series; it is part of an effort aiming at the design of derivational paradigms.
BAPs and FAPs have been adapted to the datasets of the current task, analogizing inflectional paradigms to derivational families and pairs of inflectional paradigm cells to derivational series. For instance, (\ref{ex:serie-flex-deu-1}) presents an excerpt of an inflectional series in the inflectional paradigm of the verb \emph{anspielen} that realizes the features \texttt{V;NFIN} and \texttt{V.PTCP;PST}.  In turn, this series yields two series of word-forms, the ones in the left column and the ones in the right column.
\enumsentence{
  \label{ex:serie-flex-deu-1}
  \evnup[5pt]{%
    \begin{tabular}{lll}
      \textipa{apzu\textlengthmark{}x@n} & \textipa{apg@zu\textlengthmark{}xt} & `to search'\\
      \textipa{aplOx@n}  & \textipa{apg@lOxt} & `to punch off'\\
      \textipa{aprYk@n} & \textipa{apg@rYkt} & `to disengage'\\
      \textipa{apgUk@n} & \textipa{apg@gUkt} & `to peek'
    \end{tabular}%
  }
}

\paragraph{Basic preprocessing.}

The forms in the test set (d) being in IPA, we only computed phonological BAPs and FAPs.    BAPs and FAPs have been computed for all the entries of all four datasets.  In addition, two basic modifications were performed. First, particles were reorder so as to  appear in the same position in the infinitival and inflected word-forms.  For instance, \emph{wechsele über},  \emph{\textipa{vEks@l@ y\textlengthmark{}b@r}} `to switch over' is reordered as \emph{\textipa{y\textlengthmark{}b@rvEks@l@}}. Second, all phonemes represented by digraphs and trigraphs in IPA were replaced with arbitrary unigraphs (capital letters; eg.\ \texttt{S} is substituted for \emph{\textipa{y\textlengthmark{}}}).

\paragraph{Broad alternation patterns.}

Each entry in the datasets consists of the infinitive and another form of some lexeme, accompanied by the  UT of the second form.  The BAP of a pair of forms is computed through an alignment of the two word-forms and the identification of their common parts and their differences.  The alignment is computed by means of the \texttt{SequenceMatcher} method of the python \texttt{difflib} library; we then go through the sequences provided by the method and create the word-form patterns by replacing the common parts by \texttt{+} and copying the differences in their respective patterns.  For example, \texttt{SequenceMatcher} aligns the forms \emph{\textipa{apta\textsubarch{i}l@n}} and \emph{\textipa{apg@ta\textsubarch{i}lt}} as in (\ref{ex:alignement}) which yields the \texttt{++en/+ge+t} BAP.  BAPs are therefore calculated separately for each entry considering only the two forms.
\enumsentence{
  \label{ex:alignement}
  \evnup[5pt]{%
    \begin{tabular}{l|llll}
      \hline
      word-form1 & \textipa{ap} &              & \textipa{ta\textsubarch{i}l} & \textipa{@n}\\
      word-form2 & \textipa{ap} & \textipa{g@} & \textipa{ta\textsubarch{i}l} & \textipa{t}\\
      \hline
      BAP$_1$ & + & & + & \textipa{@n}\\
      BAP$_2$ & + & \textipa{g@} & + & \textipa{t}\\
      \hline
    \end{tabular}%
  }
}

Note that a BAP can also be seen as a characterization of an analogical series.  For instance, the pairs of forms in (\ref{ex:serie-flex-deu-1}) can all be aligned in exactly the same way as in (\ref{ex:alignement}), they all have the same BAP \texttt{++\textipa{@}n/+g\textipa{@}+t} and they form formal analogies \citep{lepage1998.coling,lepage2004.coling,lepage2004.entcs,stroppa2005.conll,langlais2008.coling}.  More specifically, if two pairs of forms $(F_1,F_2)$ and $(F_3,F_4)$ have the same BAP,  then $F_1:F_2::F_3:F_4$.
BAPs could also be computed for entire inflectional paradigms as proposed by \citet{hulden2014.generalizing}.
Also note that BAPs are not specific to an inflection class, as two classes may exhibit common behavior in one part of their paradigm but not in another. For instance, the BAP \texttt{+/+s}  describes the formal relation that connects the infinitive and the \texttt{V;PRS;3;SG} form of both regular (\emph{work}) and irregular English verbs (\emph{eat}).

\paragraph{Fine alternation patterns.}

Unlike BAPs which are derived solely from the examination of pairwise alternations, FAPs rely on the place of the two word-forms in the overall morphological system to identify more stable recurrent partials corresponding to traditional exponents. For instance, the BAP relating the  German weak verbs like \emph{anspielen} to its present participle  \emph{anspielend}, relying on the optimal alignment between the two forms, does not identify the infinitive and past participle exponents \emph{\mbox{-en}} and \emph{\mbox{-end}}. These cannot be deduced from an isolate pair of word-forms, and require considering, across the paradigm,  all the pairs of word-forms that include infinitives or present participles and finding out the pair of endings that best characterizes, across lexemes, the infinitives ``similar'' to \emph{anspielen} and the present participles ``similar'' to \emph{anspielend}.  The main challenges in the identification of the FAPs are then (\emph{i}) that they involve the entire dataset and cannot be computed locally for a single pair of word-forms;  (\emph{ii}) that we need to formally define what ``similar'' means; (\emph{iii}) that we potentially need to consider all the APs of all the pairs of words included in the dataset; (\emph{iv}) that we need a reasonable operational approximation of what could considered as linguistically relevant.

\subparagraph{(\emph{i})}
The regularities that determine the FAPs are holistic properties of the dataset, i.e.\ of the union of the datasets (a), (b), (c) and (d).  The consequence is that each FAP depends on the entire dataset, and FAPs have to be recomputed each time any of the datasets (a), (b), (c) or (d) is modified.

\subparagraph{(\emph{ii})}
The pairs in (\ref{ex:serie-flex-deu-1}) are good examples of what similar may mean, from an inflectional point of view.  This type of similarity can be defined in terms of analogy. We first assume that pairs of forms that satisfy the same BAP constitute an analogical series (as they satisfy the formal analogy encoded by the BAP). Word-forms belonging to the same column in an analogical series are then considered as similar.  In our example, the word-forms in each column in (\ref{ex:serie-flex-deu-1}) count as similar.

\subparagraph{(\emph{iii})}
We limit the number of patterns to be considered by looking only at the ones that are involved in the characterization of similar word-forms.  In other words, once the sets of similar word-forms are created, we only  consider the similarities that exist between the word-forms that belong to each set, since only these ones may be part of an FAP.

Let $\Phi$ be the set of pairs of word-forms satisfying some BAP, and $\Phi_1$ (resp. $\Phi_2$) the set of word-forms that are the first (resp. second) element of a pair in $\Phi$.  What we are looking for are the patterns that characterize a large enough subset of the word-forms in $\Phi_1$ that are in correspondence with patterns that characterize a large enough subset of the word-forms in $\Phi_2$, i.e.\ such that the pair of patterns characterize  a large enough subset of the pairs in $\Phi$.  These APs are obtained as follows. 

We first collect the  WPs that possibly characterize the word-forms in $\Phi_1$ by computing a pattern of word-forms for each pair of word-forms made up of two word-forms from $\Phi_1$. These patterns are dual of the ones illustrated in (\ref{ex:alignement}) as we need to represent what the word-forms have in common and not their differences.  For instance, in the second column of (\ref{ex:patterns-1}),  the pattern that describes the common part of \emph{\textipa{apg@zu\textlengthmark{}xt}} and \emph{\textipa{apg@lOxt}} is \texttt{ap\textipa{g@}+xt} and the one for the common part of \emph{\textipa{apg@lOxt}} and  \emph{\textipa{apg@rYkt}}  is \texttt{ap\textipa{g@}+t}.  If the number of word-forms in the column is large and varied enough, all the relevant WPs that characterize a part of the word-forms will be collected.  We then align the patterns obtained for the two column.  This is done by considering the WPs as if they were word-forms and computing their analogical signature, i.e.\ their BAP.  For instance, we have \emph{\textipa{aplOx@n}\thinspace:\thinspace\textipa{aprYk@n}\thinspace::\thinspace\textipa{apg@lOxt}\thinspace:\thinspace\textipa{apg@rYkt}}.  The BAP for the first pair is \texttt{++\textipa{@n}/+\textipa{g@}+t} and the BAP for the second is identical; the pattern that characterizes \textipa{aplOx@n}:\textipa{aprYk@n} is  \texttt{ap+\textipa{@}n} and the one that characterizes the second is \texttt{ap\textipa{ge}+t}. These two WPs are aligned because their BAP is \texttt{++\textipa{@n}/+\textipa{g@}+t}, i.e. the same as the BAP of the two pairs of word-forms.

By doing the same computation for all  pairs of word-forms and matching them with respect to their BAP, we end up with a number of FAP candidates that we  first screen in order to exclude those that describe only a small part of $\Phi$, or that are made up of WPs that describe a small part of $\Phi_1$ or $\Phi_2$.  Another feature that helps select valuable FAPs is the number of variable parts (\emph{+}) it contains.  For our models, we only used FAPs that contain exactly one variable part, but this number could be increased for languages with templatic morphology or that make use of infixes.

\subparagraph{(\emph{iv})}
We  assume  that optimal FAPs  are pairs of WPs that recur both within the paradigm and across the lexicon,  as we illustrated in (\ref{ex:flex-anspielen-1}) and (\ref{ex:flex-other-verbs-1}). For instance, the FAP of a pair of word-forms \emph{\textipa{anSpi\textlengthmark{}l@n}} and \emph{\textipa{anSpi\textlengthmark{}l@nt}} consists of the aligned patterns describing  the largest number of word-forms  similar to \emph{\textipa{anSpi\textlengthmark{}l@n}} on the one hand,  and  the largest number of word-forms  similar to \emph{\textipa{anSpi\textlengthmark{}l@nt}} on the other hand.  It turns out that this is the pattern \texttt{+\textipa{@}n/+\textipa{@}nt}. 
More precisely, let $(F_1, F_2)$ be a pair of word-forms and let $\{(P_1,Q_1), (P_2,Q_2), ..., (P_n,Q_n)\}$ be the FAP candidates connecting $F_1$ and $F_2$ (i.e.\ the set of the aligned WPs of $F_1$ and $F_2$).  Let $|X|$ be the number of form pairs that satisfy an alternation that contain the WP $X$.  The  FAP of $(F_1, F_2)$ is then the WP pair $(P_i,Q_i)$ such that $|P_i|+|Q_i| = \max_{j=1}^n([P_j| + |Q_j|)$. FAPs are therefore selected separately for each pair of word-forms.

The models M1 and M2 presented in Section~\ref{sec:combining} use FAPs computed from the union of the datasets (a), (b), (c) and (d) for each of the three languages of the task.

\paragraph{Discussion.}

BAPs and FAPs give different types of information: BAPs capture relations between pairs of forms independently of the rest of the system, and are hence crucial to addressing the implicative structure of paradigms \citep{Wurzel89}. FAPs on the other hand characterize a pair of forms taking into account their place in the rest of the system; this typically leads to more specific patterns that are satisfied by a smaller set of pairs. 

\section{Combining analogical patterns and encoder-decoder networks}
\label{sec:combining}

Early work on connectionist models of the acquisition of morphology involved pattern associators that learn relationships between a base lexical form (i.e.\ the lemma) and a target form (i.e.\ the inflected form).  For example, \citet{rumelhart1986lpt} propose a simulation of how English past tense is learned. They focus on pairs of verb forms like \emph{go}-\emph{went} and \emph{walk}-\emph{walked} and consider that morphological learning is  a gradual process which includes an intermediate phases of ``over-regularization'' (where the past form of \emph{go} is \textit{goed} instead of \textit{went}). This yields the well-known``U-shape'' curves observed in the developmental phases of morphological competence in children. 

More recently, models based on deep learning architectures have been used \citep{Malouf:2017} and in particular sequence-to-sequence models able to predict one form of a lexeme from another \citep{faruqui-etal-2016-morphological,kirov-cotterell-2018-recurrent}. 

These approaches are based on the assumption that the morphological learning can reduce to a simple mapping between a base form and an inflected one.  Generalizations over similar mappings (e.g. \textit{love}-\emph{loved},\textit{walk}-\emph{walked} vs. \textit{sing}-\emph{sang}, \textit{ring}-\emph{rang}) are learned from the dependences between the phonemes in sequences.
The APs  presented in Section~\ref{sec:patterns} provide a description complementary to the lemma-form mapping in which analogical regularities may be locally to a single pair of forms (BAPs) or globally from the entire lexicon (FAPs).  These paradigmatic analogies emerge when the forms are contrasted with all other forms of their lexeme and the other forms that occupy the same cell in the paradigm \citep{BonamiBeniamine:2016,ahlberg-etal-2015-paradigm,Albright:2002}. 

The models we designed for the task combine the capacity of the sequence-to-sequence models to learn the regularities present in strings of phonemes with the alternation patterns acquired from the paradigms, in order to predict native speaker responses in a \textit{wug} test.

\subsection{Models}
\label{sec:models}

We designed four models for the shared task.  

\subsubsection{Model 1}
\label{sec:model_1}

In the first model, M1, we consider  morphological inflection as a mapping over sequences.  The mapping is implemented by bidirectional LSTMs with dropouts \citep{HochSchm97, Gal:2016}.  The hidden states of the encoder are used to initialize the decoder states.  The model adopts a teacher forcing strategy to compute the decoder's state in the next time-step.  M1 takes as input four sequences: the lemma (IPA-encoded), the UT, the BAP and the FAP patterns. The output sequence is the inflected form (IPA-encoded). The output layer uses a sigmoid  to produce a probability distribution over the output phonemes.
\begin{itemize}
  \item[M1] Input: \{lemma $+$ UT $+$ BAP $+$ FAP\} \\
  Output: \{inflected form\}
\end{itemize}

%
The probability score assigned to the \textit{wug} forms is the joint probability of the its phonemes.
The model M1 addresses the task directly.  We expected the prediction of a model that uses all the available information including BAPs and FAPs would be accurate and highly correlated with the judgments of the speakers.

\subsubsection{Model 2}
\label{sec:model_2}

The second model, M2, relies on FAPs to identify the crucial thing to be predicted in a wug task, namely the inflectional pattern of the output form. Hence the model is trained to predict, instead of the raw output form, the word pattern that constitutes the second part of the FAP (FAP$_2$) and identifies its place in the inflection system while abstracting away from what is common between the input and output forms.
\begin{itemize}
  \item[M2] Input: \{lemma $+$ UT\} \\
  Output: \{FAP$_2$\}   
\end{itemize}
Computationally M2 is similar to M1 except for the input/output sequences involved.  In particular, the probability score of a \textit{wug} form is the joint probability of the output symbols.

\subsubsection{Model 3}
\label{sec:model_3}

Our third model, M3, estimates a possible wordlikeness effects due to phonological similarity of the inflected forms that have the same UT.  Wordlikeness is the extent to which a sound sequence composing a form is phonologically acceptable in a given language.  It mostly depends on the phonotactic knowledge of the speakers \citep{vitevich:2005,Hahn_Bailey:2005} and on the existence of phonologically similar words in the mental lexicon \citep{Albright07gradientphonological}. 
For example, a \textit{wug} past form like \emph{\textipa{saIndId}} included in the English test dataset could trigger wordlikeness effect because it is similar to  an  attested past form \emph{\textipa{saIdId}} (\emph{sided}).
For each of the three languages, we designed a classifier which predicts whether an inflected form is assigned to a specific UT in the train set.  The target UTs are the ones of the inflected forms in the three test sets (d), namely\texttt{V;PST;1;SG} for ENG, \texttt{V;PST;PL} for NLD and \texttt{V.PTCP;PST} for DEU.
Technically, for each language, the M3 model is an LSTM-based binary classifier which takes the inflected form as input and outputs whether it is assigned to the target UT (value $1$) or not (value $0$).  At training time, the forms which are assigned to the target UT and to another UT, are only kept with their target UT. 
\begin{itemize}
  \item[M3] For inflected UT in the test set (d),\\Input: \{inflected form\} \\ Output: \{$0$,$1$\}
\end{itemize}
The score assigned to the \textit{wug} form is simply the probability outputted by the system. 

\subsubsection{Model 4}
\label{sec:model_4}

 Our fourth model, M4, simply uses the type frequency of the BAP relating the \textit{wug} lemma and the \textit{wug} form as a score for the test dataset.

\begin{itemize}
  \item[M4] Raw type frequency of the BAP relating the \textit{wug} lemma and \textit{wug} form  
\end{itemize}

This is meant as a very simple baseline, capturing in a very crude fashion the intuition that speakers judge as more natural wugs that fit into a more frequent pattern.

\subsection{Results}
\label{sec:results}

The submissions to the task are evaluated using the AIC scores from a mixed-effects beta regression model \citep{glmmTMB} where the scaled human ratings (DV) were predicted from the submitted model's ratings (IV). The regression implements a random intercept for lemma type.
Table \ref{AIC} reports the AIC scores of the test data for the three languages. 
\begin{table}[ht] 
\centering
\begin{tabular}{ |c|rrr| } 
 \hline
 Models & ENG & NLD & DEU\\ 
  \hline
M1 & $-33.4$ & $-60.0
$ & $-16.1$ \\ 
M2  & $\mathbf{-43.0}$ & $\mathbf{-66.0}$ & $\mathbf{-98.8}$\\ 
M3  & $-37.5$ & $-64.9$ & $-12.9$\\ 
M4  & $-40.7$ & $-36.8$ & $-72.9$ \\ 
\hline
\end{tabular}
\caption{AIC scores calculated on the basis of the final test data.  Lower scores are better.}
\label{AIC}
\end{table}

\section{Discussion}
\label{sec:discussion}

The performance of our four models suggest the following observations. First, M2 outperforms our three other models for all three languages, and also ranked second of all systems submitted to the shared task. Second, there is a striking difference in performance between M2 and M1, which had an similar architecture, but performed very poorly---worse than our baseline M4 model, and second to last of all systems submitted to the shared task. Although more experiments are needed to conclude on this point, we conjecture that the better performance of M2 might be due to the fact that it abstracts away from the question of predicting the shape of the stem in the output, but focuses instead on that part of the inflected form that is not to be found in the input. This seems to match intuitions about human behavior: when dealing with inflections, speakers may have a hard time applying the right pattern, but they never have a hard time remembering what the stem looks like, even if it is phonotactically unusual (see \citealt{Virpioja_2018} for a psycho-computational study). 

The other surprising result is that M4, which was intended as a crude baseline, did surprisingly well on the English and German data, although it performed very poorly on Dutch. This is interestingly complementary to the performance of M3, which did surprisingly well on Dutch but poorly on German. As M3 is entirely focused on phonotactic similarity while M4 is focused on the frequency of alternations, this suggests that the three inflection systems (to the extent that they are faithfully represented by the datasets) raise different kinds of challenges to speakers.

To better assess the quality of  M2, we examined how well it statistically correlates with human performance in \citeauthor{albright-hayes-2003-rules}'s (\citeyear{albright-hayes-2003-rules}) experiments on islands of reliability (IOR) in regular and irregular inflection in English. Albright and Hayes are trying to establish that speakers rely on structured linguistic knowledge as encoded in their Minimal Generalization Learner (MGL, \citealp{Albright_Hayes_2006}) rather than pure analogy when inflecting novel words. To establish this, they collected both productions of human participants asked to inflect a novel word, and judgments on pairs of word-forms. They show that the MGL leads to a better correlation with human results than a purely analogical system based on \citet{Nosofsky90} (NOS in the table below). As Table~\ref{tab:MGL} shows, our M2 performs at a level comparable to the MGL. More precisley,  it clearly outperforms it on irregular verbs while trailing  on regulars. Importantly, M2 does that without relying on any structured knowledge of the kind found in the MGL, although it does rely on a more complete view of the morphological system. This suggests that the conclusions of Albright and Hayes should be reconsidered. 

\begin{table}[ht] 
\centering

  \begin{tabular}{|l|c|c|c|c|}
    \hline
    \multirow{3}{*}{Models} &
      \multicolumn{2}{c|}{Ratings} &
      \multicolumn{2}{c|}{Production} \\
    \multirow{3}{*}{} &
      \multicolumn{2}{c|}{} &
      \multicolumn{2}{c|}{probabilities} \\
      \cline{2-5}
    & reg. & irr. & reg. & irr.  \\
    \hline
MGL    & $0.745$ & $0.570$ & $0.678$ & $0.333$  \\
NOS    & $0.448$ & $0.488$ & $0.446$ & $0.517$  \\
M2    & $0.583$  & $0.595$ & $0.611$ & $0.560$  \\
    \hline
  \end{tabular}\caption{Pearson correlations ($r$) of participant responses to models. Core IOR verbs ($n =  41$). See \citet{albright-hayes-2003-rules} for the list of nonce verbs exploited in the experiment}
\label{tab:MGL}
\end{table}

\section{Conclusion}
\label{sec:conclusion}

At the time of writing, we do not have the descriptions of the other systems that were submitted to the shared task. As a result, we are not able to identify the reasons for the good and not so good performance of the four systems we submitted.  The objective of our participation was to test different hypotheses.  The main one is the relatively low importance of stems when predicting the acceptability of wug forms, as evidenced by the good performance of the M2 model, which  which only predicts the FAP of the inflected form. Therefore, M2 is output-oriented in the sense that the properties that characterize the input, i.e. the lemma, are not used during training.

M1 and M2 models are able to predict inflected forms and FAP$_2$ patterns for any UT in the training set while M3 models are specialized on a single UT. In future work, we plan to develop specialized versions of M1 and M2 in order to estimate the importance of the tested inflectional series (i.e.\ of the set of form pairs with the same UTs as the entries in test set) with respect to the entire training set.  We further plan to test our models on more complete datasets in which the inflected forms could be predicted from other forms than the lemma, but also jointly from several forms of the lexeme \cite{BonamiBeniamine:2016}.

\section*{Acknowledgement} \label{acknowledgement}

Experiments presented in this paper were carried out  using the \href{https://osirim.irit.fr/}{OSIRIM platform}, that is administered by IRIT and supported by CNRS, the Region Occitanie, the French Government and ERDF.

\bibliographystyle{acl_natbib}
\bibliography{acl2021}

\end{document}